\documentclass[letterpaper, 10 pt, conference]{ieeeconf}
\IEEEoverridecommandlockouts
\let\labelindent\relax
\usepackage{enumitem}
\usepackage{balance}
\usepackage[font={small}]{caption}
\usepackage{subcaption}
\usepackage{array}
\usepackage{textcomp}
\usepackage{mathtools, nccmath}
\usepackage{graphicx}
\usepackage{amsfonts}
\usepackage{amsmath}
\usepackage{amssymb}
\usepackage{algorithm}
\usepackage{algorithmic}
\usepackage{hyperref}
\usepackage{tikz}
\usepackage{arydshln}
\usepackage{multirow}
\usepackage{bm}
\usepackage{epstopdf}
\usepackage{cite}
\usepackage{siunitx}

\DeclareMathOperator*{\argmax}{argmax} 

\usepackage{amsthm}

\newtheoremstyle{mystyle}
  {}
  {}
  {}
  {}
  {\bfseries}
  {.}
  { }
  {\thmname{#1}\thmnumber{ #2}\thmnote{ (#3)}}

\theoremstyle{mystyle}

\title{\LARGE \bf
Bayesian Optimization Meets Hybrid Zero Dynamics: \\Safe Parameter Learning for Bipedal Locomotion Control
\thanks{$^*$ Authors contributed equally.} 
\thanks{All authors are affiliated with Hybrid Robotics Group at the Department of Mechanical Engineering, University of California, Berkeley, USA. \{lzyang, zhongyu\_li, zengjunsjtu, koushils\}@berkeley.edu }
}

\author{Lizhi Yang*, Zhongyu Li*, Jun Zeng and Koushil Sreenath
}

\begin{document}

\maketitle
\thispagestyle{empty}
\pagestyle{empty}
\begin{abstract}
In this paper, we propose a multi-domain control parameter learning framework that combines Bayesian Optimization~(BO) and  Hybrid Zero Dynamics~(HZD) for locomotion control of bipedal robots.
We leverage BO to learn the control parameters used in the HZD-based controller. 
The learning process is firstly deployed in simulation to optimize different control parameters for a large repertoire of gaits.
Next, to tackle the discrepancy between the simulation and the real world, the learning process is applied on the physical robot to learn for corrections to the control parameters learned in simulation while also respecting a safety constraint for gait stability.
This method empowers an efficient sim-to-real transition with a small number of samples in the real world, and does not require a valid controller to initialize the training in simulation. 
Our proposed learning framework is experimentally deployed and validated on a bipedal robot Cassie to perform versatile locomotion skills with improved performance on smoothness of walking gaits and reduction of steady-state tracking errors.

\end{abstract}

\section{Introduction}
Bipedal robot locomotion is an active area of research with several control design challenges in order to deal with its high degrees-of-freedom (DoFs) and hybrid nonlinear dynamics.
Despite the control complexity, a bipedal robot is a versatile option in many application areas. 
To address this control complexity, Hybrid Zero Dynamics~(HZD)~\cite{GrChAmSi2010} was introduced to stabilize energy-efficient periodic gaits by using nonlinear controllers on the full-order robot model.
However, the performance of the HZD-based controller highly depends on the choice of the various controller gain parameters used. 
For high-dimensional robots like Cassie, the complexity of designing such a controller  
could easily make it hard to tune by hand especially when the controller parameters need to be scheduled based on different gaits~\cite{li2020animated}.
\begin{figure}[!tp]
    \centering
    \includegraphics[width=\linewidth]{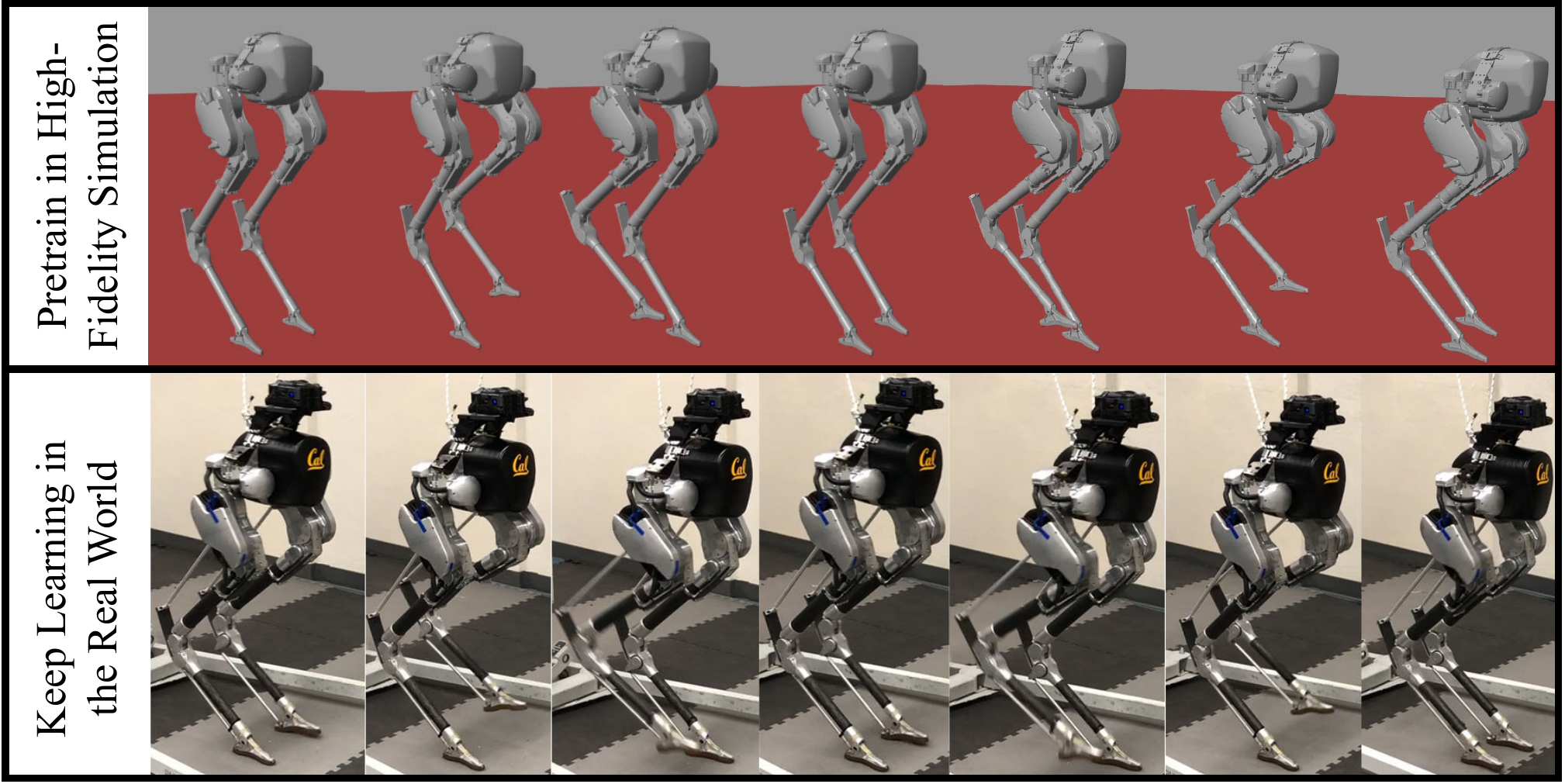}
    \caption{The proposed control parameter learning framework that applies Bayesian optimization to a HZD-based controller on a bipedal robot Cassie. Controller parameters are firstly learned in a high fidelity dynamic simulator. 
    The proposed learning process is then applied on the physical robot to learn corrections to the control parameters learned in simulation. 
    The robot successfully learns to maintain stable gaits in the real world, see experiment video at \url{https://youtu.be/WxkdJdMRdfM}.
    }
    \label{fig:main}
    \vspace{-0.35cm}
\end{figure}

When the full-order dynamics of a bipedal robot is considered as a black-box function with measurable inputs and outputs, data-driven optimization schemes, such as Bayesian optimization~(BO)~\cite{berkenkamp2021bayesian} can be applied to find the optimal control policy.
Compared to reinforcement learning~(RL) methods that are data hungry~\cite{li2021reinforcement,xie2020learning,siekmann2021sim,castillo2021robust}, BO, is a model-based method, is sample efficient and has the ability to impose additional safety constraints to provide formal guarantees of safety.
The objective of this work is to provide a safe and practical scheme to enhance HZD-based bipedal walking controllers by learning optimal control parameters via BO. 
Such a learning framework utilizes BO on multiple domains.
We first leverage a high-fidelity dynamics simulation domain to learn the initial control parameters and prevent the random initialization from destabilizing the physical robot, potentially damaging itself or harming nearby humans in the real world.
Next, we apply this framework on to the physical robot to learn the discrepancy between simulation and practice while ensuring safety constraints found in the simulated domain for gait stability, as shown in Fig.~\ref{fig:main}.

\subsection{Related Work}
\subsubsection{HZD-based Bipedal Locomotion Control}
Hybrid Zero Dynamics is a nonlinear control technique to generate stable periodic locomotion behaviors on legged robots. 
A typical HZD-based control design for 3D bipedal robots is comprised of two main parts: a gait library optimized offline using full order dynamics to provide reference gaits online~\cite{hereid2017frost, hereid2018rapid}, and online gait regulators, usually PD-based, to provide corrections to the reference gait to track given control commands~\cite{da20162d,gong2019feedback,li2020animated,reher2021control,WAFR2016GaitLibrarySteppingStones,RSS2017DiscreteTerrainWalking}. 
Among these, a controller that is able to significantly change the robot's walking height is developed in~\cite{li2020animated} based on~\cite{gong2019feedback}, by including a gain scheduling module to adjust the PD gains with respect to the commanded walking height.
However, such a gain scheduling approach on a complex dynamically coupled system that couples velocities and walking heights requires lots of tedious tuning both in simulation and in the real world. Even if a set of gains is successfully tuned by hand, it may still be sub-optimal in terms of control performance such as steady state errors and oscillations.  
Such a challenge is not only faced by~\cite{li2020animated} but is an open problem for the controllers using HZD, such as~\cite{gong2019feedback,reher2021control}. 
Therefore, in this work, we propose to leverage BO to learn optimal control gains in HZD-based controllers for bipedal robots.


\subsubsection{Bayesian Optimization in Robotics}
In Bayesian optimization,
an approximate model of the objective function is actively learned by making regular assumptions of the mean and variance of the Gaussian Process (GP)\cite{rasmussen2003gaussian,DBLP:conf/icml/SrinivasKKS10,bull2011convergence} instead of treating it as a black box, enabling it to estimate the global optima accurately with few evaluations by guiding the choice of inputs aimed at locating the most informative points~\cite{jones2001taxonomy,mockus2012bayesian}. 
BO was applied in the legged robotics community to perform quadrupedal robot gait~(periodic joint trajectory) optimization in~\cite{lizotte2007automatic} and for 
the forward walking speed of a small bipedal robot in ~\cite{calandra2016bayesian}.
Recently, user-preference-based BO gait optimization is introduced on a lower-limb exoskeleton~\cite{tucker2020preference}.
However, when such a method is extended to bipedal robots~\cite{csomay2021learning}, it only exemplifies the ability to optimize for a single walking gait on a treadmill.
As we will see, instead of applying BO on gait optimization, our method applies BO for learning the high-dimensional control parameters for Cassie. 
We are able to learn control parameters for a diverse range of gaits while only requiring small number of samples in the real world.

Furthermore, while applications of BO in robotics~\cite{tesch2011using,ryou2021multi} showcase its potential, safety in the learning process in the physical world is not generally considered as a critical process. 
Safety in terms of gait stability is critical for person-sized bipedal robots like Cassie.  
Algorithms to find the optima without violating safety constraints during BO were studied in \cite{10.5555/3020751.3020778,sui2015safe,schreiter2015safe}. 
Research in \cite{berkenkamp2016safe} introduces hard safety constraints to the optimization process, and work in \cite{berkenkamp2021bayesian} furthers it by introducing multiple constraints.
Research in \cite{marco2021robot} deploys BO on a jumping quadruped with motor current constraints, but no dynamic constraints. 
In this work, we constrain our evaluation within a safe set for gait stability of the bipedal robot Cassie during the real-world experiments to prevent the BO process from causing the robot to fall over.

\subsection{Contributions}
The contributions of this work are:
(1) We apply Bayesian optimization to a HZD-based control framework to learn optimal control parameters and improve the performance of such a controller.
(2) We develop a control parameter learning framework based on BO for HZD-based bipedal walking control on person-sized $3D$ bipedal robots in multiple domains.
This framework leverages a high-fidelity dynamics simulator to provide a safe environment to learn for control parameters for different gaits, and to obtain a safety constraint for the robot to maintain gait stability that can be utilized in the fine tuning process in the real-world experiments.
(3) The proposed framework is experimentally deployed and validated on a bipedal robot Cassie to perform versatile locomotion skills including walking with varying saggital and lateral speeds, and changing and maintaining different walking heights. 
The introduction of the proposed method brings clear improvements over previous HZD-based control implementation in terms of control performance such as steady-state tracking errors and smoothness of walking gaits with a small number of samples.

\section{Cassie and HZD-based Walking Control}\label{sec:cassie-hzd}

\subsection{Cassie Model}
Cassie is a dynamic bipedal walker 
with 20 DoFs, captured by the generalized coordinates $\mathbf{q} \in \mathbb{R}^{20}$, as shown in Fig.~\ref{fig:main}. 
There are 2 unactuated joints and 5 actuated joints $\mathbf{q}_m = q_{1,2,3,4,7}^{L/R}$  on its Left/Right legs, and its pelvis, which is a floating base, has 6 DoFs denoted by $q_{x,y,z,\psi,\theta,\phi}$ representing translational, sagittal, lateral and vertical positions, and rotational roll, pitch and yaw positions, respectively. 
A detailed introduction of Cassie can be found in~\cite{li2020animated}.

\subsection{Variable Walking Height Control via HZD}
\begin{figure}[!]
    \centering
    \includegraphics[width=0.9\linewidth]{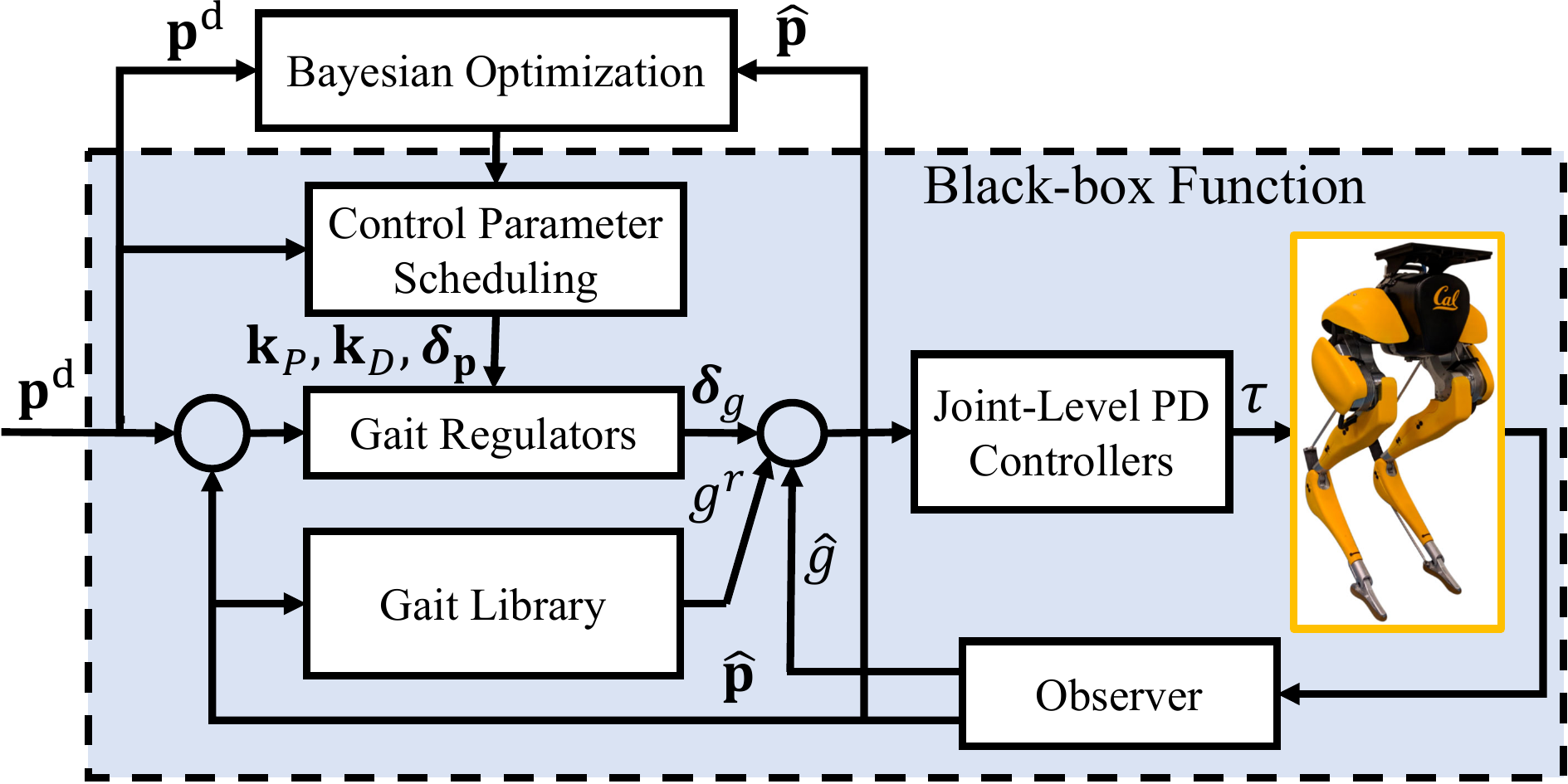}
    \caption{ The proposed BO framework is applied on the HZD-based variable walking height controller for Cassie that is previously developed in~\cite{li2020animated} based on~\cite{gong2019feedback}. This controller linearly interpolates a gait library with respect to observed gait parameter $\hat{\mathbf{p}}$ to obtain a reference gait $g^r$. PD-based gait regulators provide correction $\bm{\delta}_{g}$ to $g^r$ to track the desired gait parameter $\mathbf{p}^\text{d}$. A joint-level controller is used to generate motor torques $\tau$ based on the desired gait $g^r + \bm{\delta}_{g}$ and observed gait $\hat{g}$. The controller parameters including the PD gains $\mathbf{k}_P$, $\mathbf{k}_D$ and command tracking offset $\bm{\delta}_\mathbf{p}$ used in the controller are scheduled in a look-up table. This control parameter table is obtained by the proposed BO framework by considering the closed-loop feedback system as a black-box function.}
    \label{fig:controller}
    \vspace{-0.3cm}
\end{figure}

\subsubsection{HZD-based controller}
In this work, the walking controller for the Cassie is extended from a variable walking height controller proposed in \cite[Sec. IV]{li2020animated} based on~\cite{gong2019feedback}. 
This controller, as illustrated inside the dashed box in Fig.~\ref{fig:controller}, is comprised of two main parts: a gait library optimized offline to provide a reference gait online, and PD-based online gait regulators to provide corrections to the reference gait to track given control commands.
 
In this controller, we obtained 1331 dynamically-feasible walking \textit{gaits} $g_{\mathbf{p}}$ for different gait parameters $\mathbf{p}=[\dot{q}_x, \dot{q}_y, q_z]^T$ offline using the robot's hybrid full-order model and HZD~\cite{hereid2017frost,hereid2018rapid}.
A \textit{gait} $g_{\mathbf{p}}$ is defined as a set of periodic trajectory of joints represented by B\'ezier curves and is parameterized by $\mathbf{p}$, and we formulate these different gaits as a \textit{gait library} $\mathcal{G} = \{g_{\mathbf{p}}\}$. 
We encourage readers to refer to~\cite[Sec. IIIB]{li2020animated} for more details of such a gait library. 

In order to control the robot's locomotion online, gaits were altered by interpolating neighboring pre-computed gaits retrieved from the gait library with respect to the robot's current gait parameters $\hat{\mathbf{p}}$. 
The regulating term $\bm{\delta}_{g}$ can be obtained by
\begin{equation}
    \begin{aligned}\label{eq:controller}
    \bm{\delta}_{g}=\mathbf{k}_P(\mathbf{p}^\text{d})(\mathbf{p}^\text{d}+\bm{\delta}_{\mathbf{p}}(\mathbf{p}^\text{d})-\hat{\mathbf{p}})+\mathbf{k}_D(\mathbf{p}^\text{d})(\Dot{\mathbf{p}}^\text{d}-\Dot{\hat{\mathbf{p}}}).
    \end{aligned}
\end{equation}
The desired motor position $\mathbf{q}^\text{d}_m$ then is the sum of the reference gait $g^r$ and regulating term $\bm{\delta}_{g}$ at the current time. 
Moreover, $\bm{\delta}_{\mathbf{p}}$ presents command tracking offsets which is introduced to mitigate the steady state tracking error of such PD-based gait regulators.

This walking controller has a control parameter scheduling module where the aforementioned PD gains $\mathbf{k}_P \in \mathbb{R}^3$, $\mathbf{k}_D \in \mathbb{R}^3$ and command tracking error $\bm{\delta}_{\mathbf{p}} \in \mathbb{R}^3$ are scheduled by linearly interpolating in a look-up control parameter table with respect to desired gait parameter $\mathbf{p}^\text{d}$. We denote $\mathbf{k}=[\mathbf{k}_P^T, \mathbf{k}_D^T]^T$ and $\bm{\delta}_{\mathbf{p}}$ as \textit{control parameters}. In this way, the controller enables Cassie to track desired saggital velocity $\dot{q}^\text{d}_x$, lateral velocity $\dot{q}^\text{d}_y$, and walking height $q^\text{d}_z$ with the control parameters scheduled with respect to different gaits. Details of this controller can be found in~\cite[Sec.IV]{li2020animated}.



If each different gait parameter $p$ can have well-tuned control parameters ($\mathbf{k}$, $\bm{\delta}_\mathbf{p}$), the controller performance can be improved.
This is motivated by previous control design experiences where the control parameters working well in the forward gaits may not be the best for the backward ones.
Therefore, as shown in Fig.~\ref{fig:controller}, Bayesian optimization is introduced to learn the optimal control parameters, $\mathbf{k}$, and $\bm{\delta}_{\mathbf{p}}$, to be used in the control parameters scheduling module. 
\section{Control Parameter Learning via Bayesian Optimization}
\label{sec:bayesian-opti}

In this section, we present details of how to implement multi-domain Bayesian optimization on Cassie to achieve a safe control parameter learning procedure. The goal is to optimize different sets of control parameters for different combinations of gait parameters $\mathbf{p}$.

\subsection{Problem Formulation}
\begin{figure}
    \centering
    \includegraphics[width=0.9\linewidth]{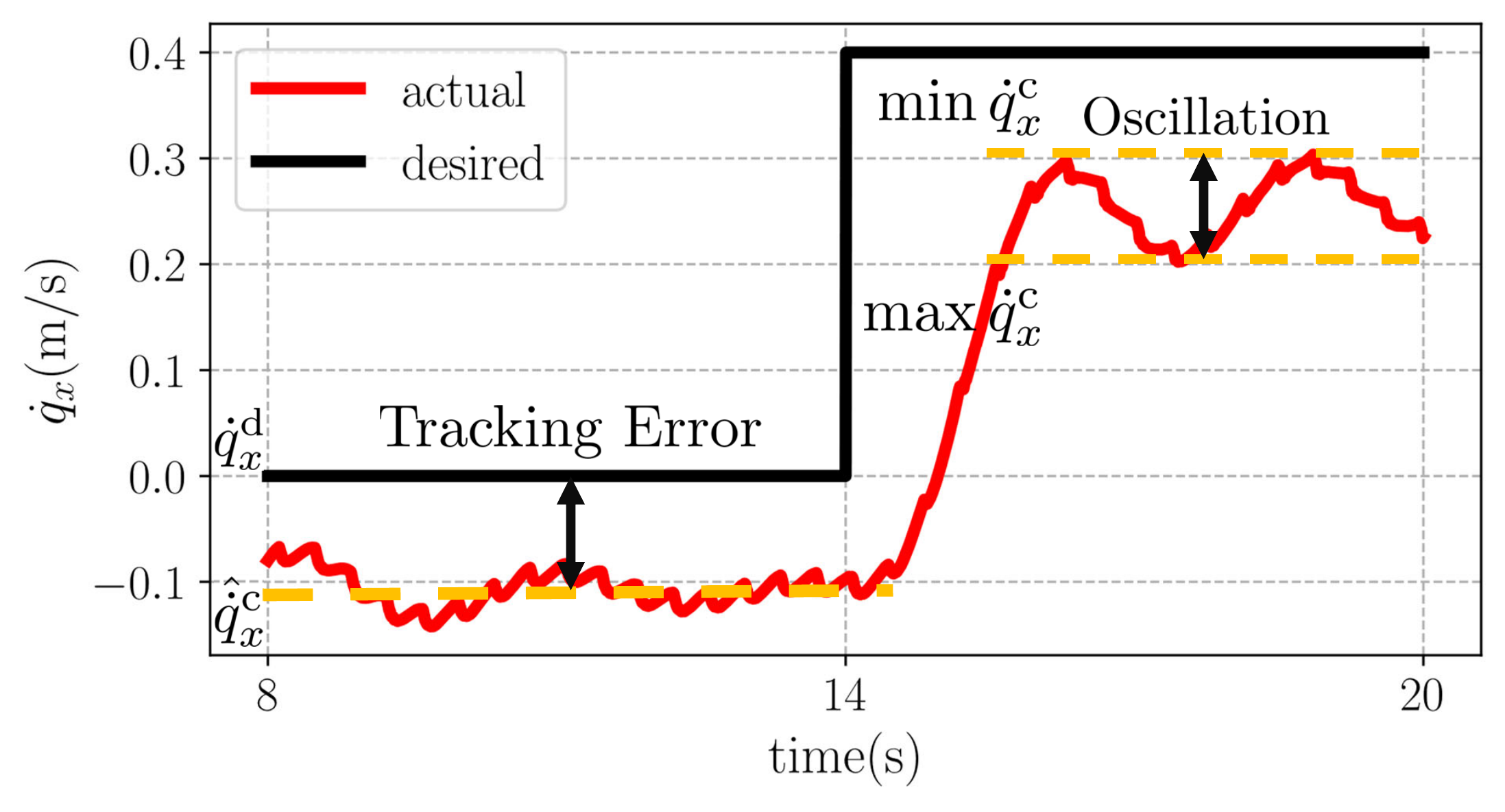}
    \caption{An example of a step response of the sagittal walking speed $\dot{q}_x$ of Cassie controlled by its HZD-based controller. The objective function is defined to minimize the deviation of the converged observed gait parameter $\hat{\mathbf{p}}^\text{c}$ with respect to the desired one $\mathbf{p}^\text{d}$ and the oscillation magnitude after convergence is defined by $||\hat{\mathbf{p}}^{\text{c}}_{\text{max}} - \hat{\mathbf{p}}^{\text{c}}_{\text{min}}||$.}
    \label{fig:cost-function}
    \vspace{-0.4cm}
\end{figure}
Our proposed algorithm uses BO to find appropriate control parameters to efficiently minimize tracking errors and oscillations.
The objective function is defined as follows,
\begin{equation}
\label{eq:objective-function}
   \min_{\mathbf{k}\in \mathbb{R}^6, \bm{\delta}_{\mathbf{p}} \in \mathbb{R}^3} f(\mathbf{k}, \bm{\delta}_{\mathbf{p}}) = 
    ||\hat{\mathbf{p}}^{\text{c}} - \mathbf{p}^\text{d}||_{\mathbf{w}_1}^2 +
    ||{\mathbf{p}}^{\text{c}}_{\text{max}} - {\mathbf{p}}^{\text{c}}_{\text{min}}||_{\mathbf{w}_2}^2
\end{equation}
where $\mathbf{p}^\text{d}$ is the desired gait parameter on which $\mathbf{k}$ and $\bm{\delta}_{\mathbf{p}}$ depend, $\hat{\mathbf{p}}^{\text{c}}$ is the converged gait parameter calculated from the average of a last segment of the sampled trajectory, and ${\mathbf{p}}^{\text{c}}_{\text{min}}, {\mathbf{p}}^{\text{c}}_{\text{max}}$ represent the range of gait parameters calculated over the last segment which lasts 5 seconds in this work. In \eqref{eq:objective-function}, the first term is chosen to minimize the deviation of the converged gait parameter with respect to the desired one, while the second term minimizes the oscillation over a duration.
The weights $\mathbf{w}_1 \in \mathbb{R}^{3\times 3}$ and $\mathbf{w}_2 \in \mathbb{R}^{3\times3}$ could be chosen to prioritize which component to focus on when minimizing both tracking error and oscillation.
Fig. \ref{fig:cost-function} visualizes the cost function for the gait parameter $\Dot{q}_x^\text{d}=0$ for $t\in[8,14]$ and  $\Dot{q}_x^\text{d}=0.4$ for $t\in(14,20]$.

Each evaluation is selected based on the acquisition function, which considers exploration and exploitation with the goal of maximizing the efficiency of the overall optimization process.
The data point is updated with the following subproblem at the $i$-th iteration, 
\begin{equation}
        \left \{
  \begin{aligned}
    \mathbf{k}_i&=\argmax_{\mathbf{k}} ~~a(\mathbf{k}|\mathcal{D}), && \text{if simulation}  \\
    \bm{\delta}_{\mathbf{k},i}, \bm{\delta}_{\mathbf{p},i}&=\argmax_{\bm{\delta}_{\mathbf{k}},\bm{\delta}_{\mathbf{p}}} ~~a(\bm{\delta}_{\mathbf{k}},\bm{\delta}_{\mathbf{p}}|\mathcal{D},\mathcal{C},\mathbf{k}), && \text{if real-world}
  \end{aligned} \right.
\end{equation}
where $a(\cdot)$ is the acquisition function, $\mathcal{D} = \mathcal{D}_1 \cup \mathcal{D}_2 ... \cup \mathcal{D}_i$ is the data set of all past evaluations with $\mathcal{D}_i = f(\mathbf{k}_i, \bm{\delta}_{\mathbf{k},i}, \bm{\delta}_{\mathbf{p},i}, \mathbf{p}_i)$, where $\mathbf{p}_i$ denotes the desired gait parameters, $\bm{\delta}_{\mathbf{p},i}$ denotes the command tracking offset, $\bm{\delta}_{\mathbf{k},i}$ denotes the corrections to be learned given a set of simulation trained PD gains in the real world, and $\mathcal{C}$ is the safety constraint to prevent the robot from losing gait stability, see Sec. \ref{subsec:multi-domain}.
In this paper, the acquisition function $a(\cdot)$ is chosen under the form of expected improvement \cite{frazier2018tutorial}. 
When the standard deviation of the posterior of the objective function is too small, we also modify the kernel function by scaling the standard deviation with an adaptive ratio to avoid over-exploiting, see \cite{bull2011convergence}.
Thus with BO, a set of control parameters can be retrieved from the objective function and the underlying dynamic model. 
Next, we present this method via a multi-domain iterative optimization approach.

\subsection{Iterative Optimization over Multiple Domains}
\label{subsec:multi-domain}
\begin{figure}[!tp]
    \centering
    \includegraphics[width=0.8\linewidth]{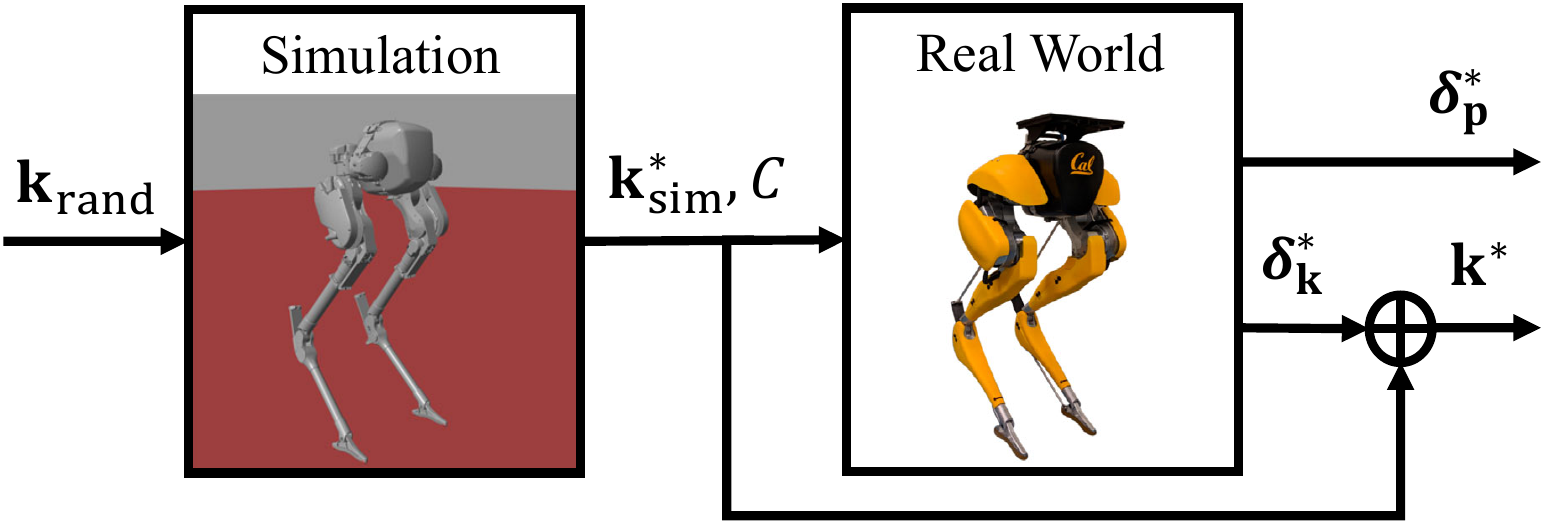}
    \caption{Overview of the proposed multi-domain control parameter learning via BO. The PD gains $\mathbf{k}$ are initialized randomly for the simulation domain and the result of the BO in simulation $\mathbf{k}^{*}_{\text{sim}}$ is used to initialize real-world learning. The correction term to the PD gain $\bm{\delta}_{\mathbf{k}}^{*}$ and command tracking offset $\bm{\delta}_{\mathbf{p}}^{*}$ are optimized using BO in the real-world experiments. Moreover, the simulation also provides the safety constraint $\mathcal{C}$ for gait stability in the real world.}
    \label{fig:system}
    \vspace{-0.3cm}
\end{figure}

The aforementioned multi-domain control parameter learning approach is applied on two domains: (a) a high-fidelity simulation in MATLAB SimMechanics where the PD gains $\mathbf{k}$ are learned, and (b) the real world where the corrections for PD gains $\bm{\delta}_{\mathbf{k}}$ and command tracking offset $\bm{\delta}_{\mathbf{p}}$ are optimized, as illustrated in Fig.~\ref{fig:system}.
The control parameters are learned for 310 different gait parameters which are separated into three sets $\mathcal{P}_{\text{sim},1,2}$ and $\mathcal{P}_{\text{real}}$ to increase the granularity of the control parameter scheduler in \cite{li2020animated} and are defined as:
\begin{align}
    \mathcal{P}_{\text{sim},1}&=\{\mathbf{p}|\mathbf{p}=[0,0,\{1,0.9,\dots,0.7\}]^T\}, \nonumber\\
    \mathcal{P}_{\text{sim},2}&=\{\mathbf{p}|\mathbf{p}=[\{-1,-0.8,\dots,0.8,1\}, \nonumber\\ &\{-0.3,-0.2,\dots,0.2,0.3\},\{1,0.9,\dots,0.7\}]^T\}\setminus \mathcal{P}_{\text{sim},1}, \nonumber \\
    \mathcal{P}_{\text{real}}&=\{\mathbf{p}|\mathbf{p}=[0,0,\{1,0.9,0.8\}]^T\}. \nonumber
\end{align}

Specifically, we firstly obtain a set of optimal PD gains $\mathbf{k}^*_{\mathcal{P}_{\text{sim},1}}$ from random initialization in the simulation domain without access to a valid initial controller for four nominal gait parameters $\mathcal{P}_{\text{sim},1}$ which are stepping-in-place gaits with different walking heights, with $I_1$ number of iteration for each.
Since the initialization is random, such a learning process requires relatively large samples empirically with $I_1=100$.
Later, having initialized the control parameter for the nominal gaits, the control parameters with all other gait parameter $\mathcal{P}_{\text{sim},2}$ are optimized next each with $I_2=25$.
Therefore, the optimal PD gains in simulation $\mathbf{k}^*_{\text{sim}}$ will be the combination of $\mathbf{k}^*_{\mathcal{P}_{\text{sim},1}} \cup \mathbf{k}^*_{\mathcal{P}_{\text{sim},2}}$.
After $\mathbf{k}^*_{\text{sim}}$ is obtained, it is utilized to generate safety constraint $\mathcal{C}$ for gait stability for the robot, and leveraged as a set of initial PD gains in the real-world experiments. 
To deal with the sim-to-real gap, the corrections to the PD gains $\bm{\delta}_{\mathbf{k}}$ and command tracking error $\bm{\delta}_{\mathbf{p}}$ are learned during the experiments while respecting the safety constraints for three different nominal gait parameters $\mathcal{P}_{\text{real}}$ with only $I_3=10$ iterations to obtain a set of optimal corrections for each gait parameter.
The details of the multi-domain learning method are discussed as follows.

\subsubsection{Learning in Simulation}
Leveraging the benefit of a simulation, we initialize the gains $\mathbf{k}_{\text{rand}}=[\mathbf{k}_P^T,\mathbf{k}_D^T]^T$ randomly without a valid controller within reasonable bounds based on \cite{li2020animated}
and perform separate optimizations for different gait parameter set $\mathcal{P}_{\text{sim},1,2}$ in simulation over iteration $I_{1,2}$.
During each iteration, the smallest feasible mean of the posterior distribution is estimated by sampling feasible points using the acquisition function $a(\mathbf{k}|\mathcal{D}^{\text{sim}})$ and local search improvement.
The performance of $\mathbf{k}_i$ is then evaluated for the desired gait parameter $\mathbf{p}$ using the objective function $f(\mathbf{k}_i,\mathbf{p})$. 
The resulting data $\mathcal{D}_i^{\text{sim}}$ is then used to update the GP, iterative improvement it and minimizing the objective function to search for the optimal gains.
At this fidelity level, the GP model is able to incorporate stochastic measurement and actuation noise, robot dynamics and physical effects. 

\subsubsection{Safe Set for Gait Stability}
After obtaining a set of optimal controller parameters in simulation, extensive tests are applied using the resulting controller to let the bipedal robot Cassie track different gait parameters in the simulation.
During the test, every desired gait parameter $\mathbf{p}^\text{d}$ in a possible range is given to the controller, and if the robot can maintain a stable walking gait, the desired gait parameter $\mathbf{p}^\text{d}$ is recorded as part of the \emph{feasible command set}, and the average converged observed gait parameter on the robot $\hat{\mathbf{p}}$ is recorded as part of the \emph{safe set}.
The concept of these two sets are introduced in detail in~\cite{li2021reinforcement}.
A convex subset of the safe set can be obtained by choosing vertices manually to form a convex polyhedron as shown by the red region in Fig.~\ref{fig:safeset}.  
Such a \emph{convex safe set} can be utilized as a safety constraint during the learning process in the real world to prevent the robot from losing gait stability.

\subsubsection{Real-world Optimization with Safety Guarantee}
\begin{figure}[!tp]
    \centering
    \includegraphics[width=0.75\linewidth]{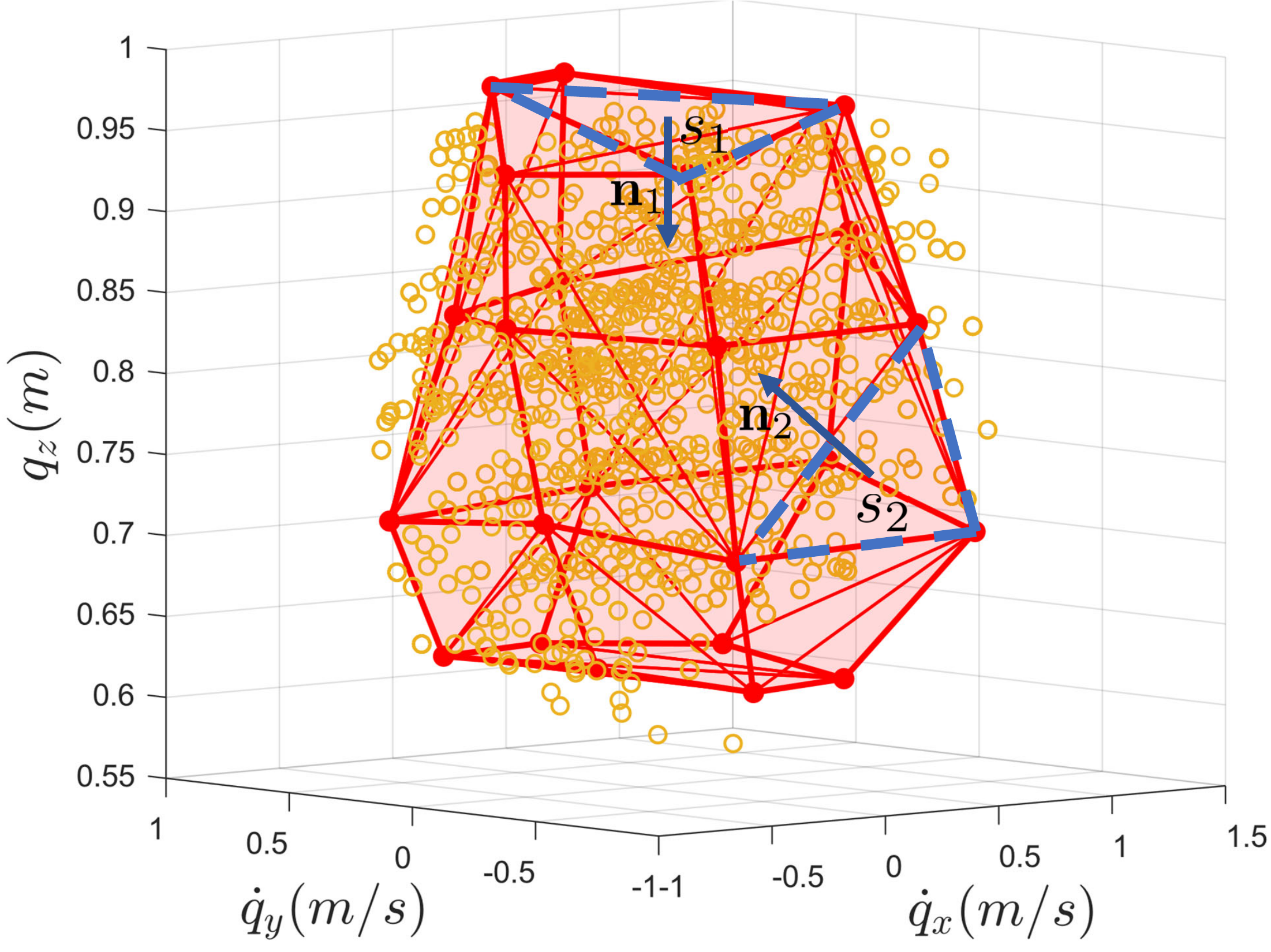}
    \caption{Safe set of gait parameters for the HZD controller.  The orange dots denote the ``safe'' gait parameters that the controller 
    exhibits in simulation for which the robot doesn't fall even after several seconds, thus denoting 
    gaits for which the controller works stably. 
    The solid red points $\mathbf{p}_{safe}$ are manually selected as vertices to formulate a convex hull whose surfaces are shaded in red. The blue lines define the surface normal $\mathbf{n}_j$ of a chosen surface $s_j$.}
    \label{fig:safeset}
    \vspace{-0.5cm}
\end{figure}

After a set of gains $\mathbf{k}^{*}_{\text{sim}}$ has been optimized, we next perform BO on the real-world robot. This enables us to find control parameters that are suitable for the true but unmodeled dynamics of the full system, including motors dynamics, vibration modes, and battery dynamics.
In this real-world domain, correction terms for the PD gain $\bm{\delta}_{\mathbf{k}} = [\delta k_P^{\dot{q}_x},\delta k_D^{\dot{q}_x},\delta k_P^{\dot{q}_y},\delta k_D^{\dot{q}_y},\mathbf{0}]^T$ as well as command tracking offset $\bm{\delta}_{\mathbf{p}}=[\delta \dot{q}_x,\delta \dot{q}_y, \delta \dot{q}_z=0]^T$ need to be optimized to ensure the robot is able to track the desired gait parameters properly.
Therefore, we further run the learning process on the real-world robot, utilizing a sparser gait parameter set $\mathcal{P}_{\text{real}}$ and smaller iteration $I_3$.
After a set of corrections for PD gains is optimized, 
the optimal gains from simulation are modified such that the real-world optimal gains are $\mathbf{k}^*=\mathbf{k}_{\text{sim}}^*+\bm{\delta_{\mathbf{k}}^*}$. Along with optimal $\bm{\delta}_{\mathbf{p}}^*$, these learned parameters are used in the controller via~\eqref{eq:controller}.

During this process, safety constraints are also imposed on the robot in order to prevent the BO from picking a control parameter that causes the robot to lose gait stability.
Unlike previous work that imposes a similar constraint with nonlinear programming~\cite{li2021vision}, this is achieved by imposing coupled constraints on the GP by defining a latent constraint function $h$ such that the constraint $\mathcal{C}$ is satisfied when $h\leq 0$. 
A feasibility model of the constraint can thus be constructed by computing the posterior distribution of $h$ and setting $\text{Pr}(\mathcal{C})=\text{Pr}(h\leq 0)$ using the Gaussian CDF of the GP, and the BO problem is constrained by $\text{Pr}(\mathcal{C})\geq 1-\text{tol}$, where $\text{tol}$ is the tolerance of the constraint \cite{10.5555/3020751.3020778}. 
The safety constraints confine the gait parameters to be within the convex safe set, as illustrated in Fig.~\ref{fig:safeset}, by using an inner product to ensure that the point $\hat{\mathbf{p}}^{\text{c}}$ lies within the feasible region, such that
\begin{align}
    <\mathbf{p}_{\text{safe},j} - \hat{\mathbf{p}}^{\text{c}}, \mathbf{n}_{j}> \leq 0, \forall s_j\in S.
\end{align}
Here $S$ is the set of surfaces of the convex polyhedron that is formed by connecting vertices of the convex safe set, $\hat{\mathbf{p}}^{\text{c}}$ is the converged gait during the experiment, $\mathbf{p}_{\text{safe},j}$ are the vertices of the convex safe set~(red points in Fig.~\ref{fig:safeset}) corresponding to the surface $s_j$, and $\mathbf{n}_j$ is the inward-facing surface normal of that surface $s_j$. This ensures that only gait parameters within the polyhedron formed by the convex safe set in Fig.~\ref{fig:safeset} satisfy the constraint imposed upon the optimization process, thus ensuring the safety for the gait stability of the robot.

\begin{figure}
    \centering
    \begin{subfigure}[t]{0.99\linewidth}
        \centering
        \includegraphics[width = 0.99\linewidth]{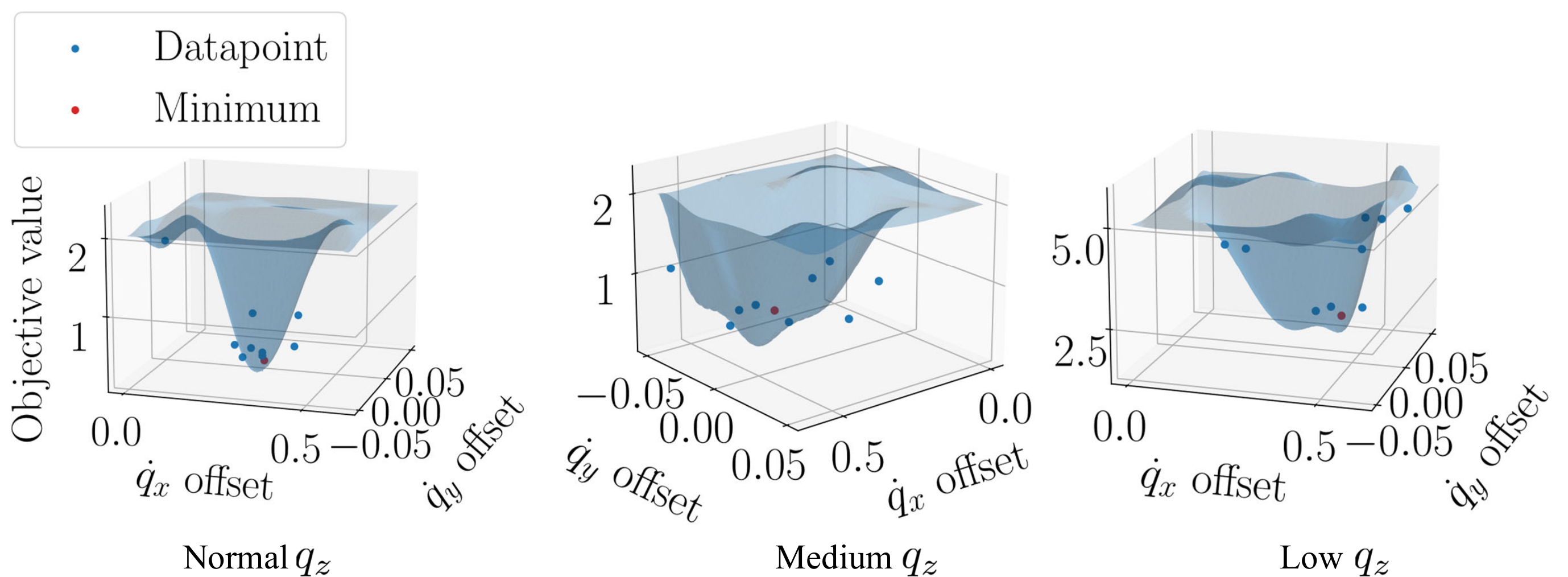}\\
        \caption{Projection of the sampled GP onto the planar velocity tracking offset $\delta \dot{q}_x$-$\delta \dot{q}_y$ space with the blue surface describing an estimation of the underlying objective function. The blue points are the data points in each iteration while the red point is the output from BO.}
        \label{subfig:3d_obj}
    \end{subfigure} \\
    \begin{subfigure}[t]{0.99\linewidth}
        \centering
        \includegraphics[width = 0.99\linewidth]{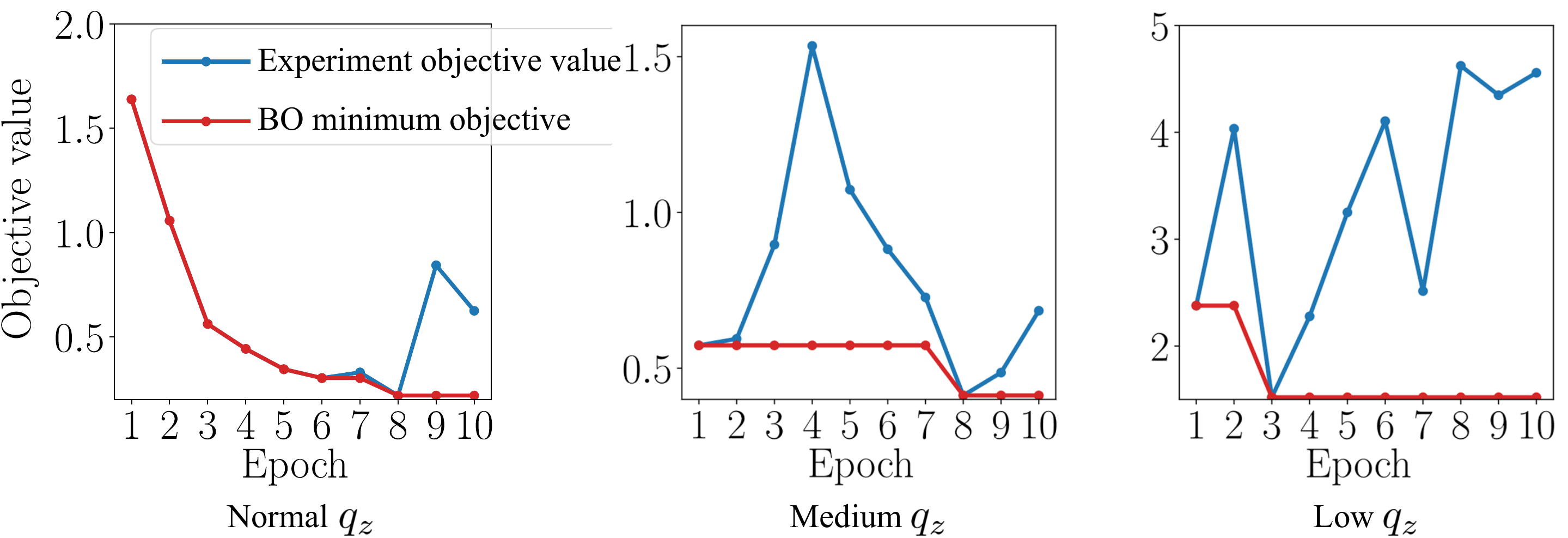}\\
        \caption{BO minimum objective values with respect to number of epochs for the 3 different walking heights are plotted in red. Experiment objective values are also plotted in blue.} 
        \label{subfig:2d_obj}
    \end{subfigure}
    \caption{Learning process in the real-world domain for three different walking heights: normal $q_z$~(1.0m), medium $q_z$~(0.9m) and low $q_z$~(0.8m). (a) Plots of the GP projection onto the command tracking offset plane. (b) BO minimum and experimental objective values with respect to number of epochs.}\label{fig:learning-process-real}
    \vspace{-0.5cm}
\end{figure}

The learning process in the real-world domain is visualized in Fig.~\ref{fig:learning-process-real}. 
As exemplified in Fig.~\ref{subfig:3d_obj}, the output from BO is always at or near to the global minimum of the estimated black-box function $f$.

\section{Experiments and Evaluations}\label{sec:experiment}
Having obtained the optimal controller parameters for the HZD-based variable walking height controller via BO, we extensively test its control performance and benchmark with a previous controller that was developed without using BO in~\cite{li2020animated} and based on a fixed walking height controller~\cite{gong2019feedback}.
The tests are launched on a bipedal robot Cassie in the simulation and real-world domains.

\begin{figure*}
\centering
\begin{subfigure}[t]{0.46\linewidth}
    \centering
     \includegraphics[width=\linewidth]{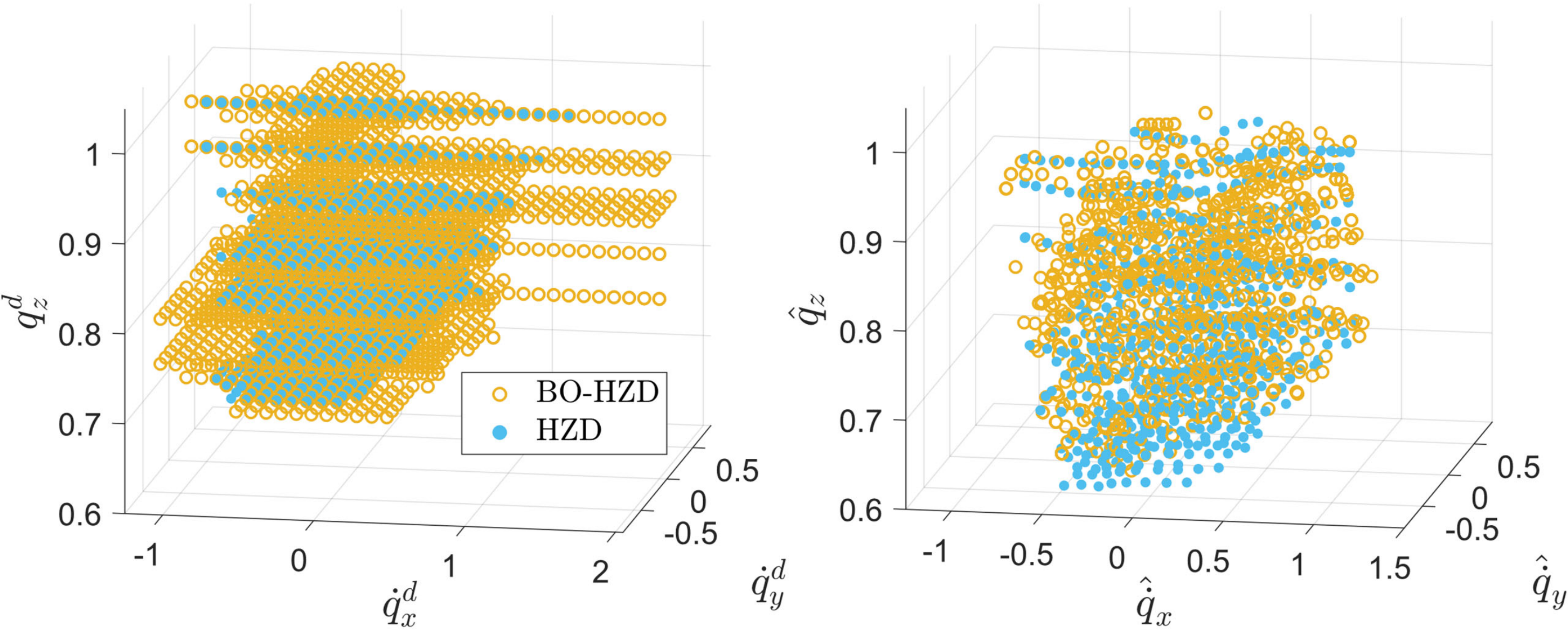}
     \caption{Comparison for feasible command set (left) and safe set (right) between the proposed BO-HZD and HZD. The BO-HZD controller has a larger command coverage and a comparable safe set.}
     \label{subfig:set-comp}
\end{subfigure}
~\begin{subfigure}[t]{0.45\linewidth}
  \centering
        \includegraphics[width=\linewidth]{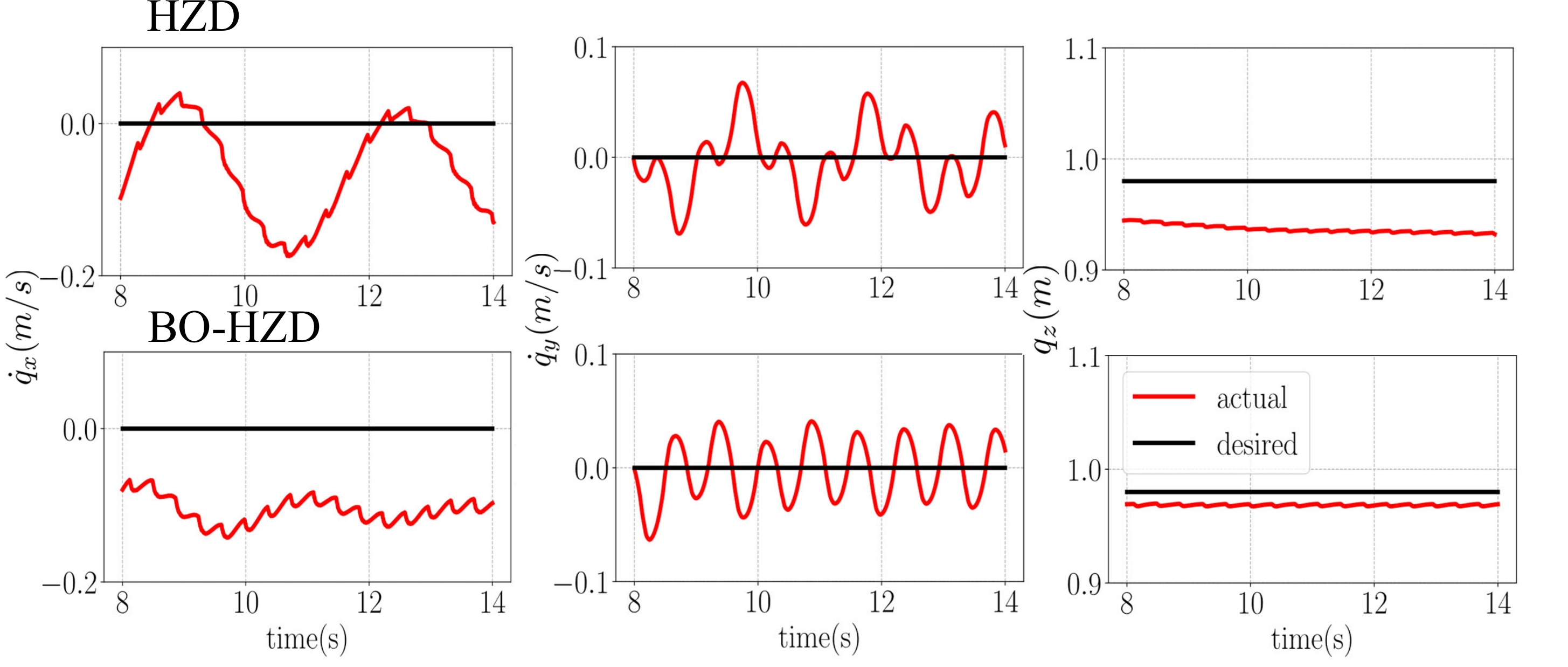}
        \caption{Comparison for controller performance between the proposed BO-HZD and HZD. The BO-HZD controller shows less tracking error and smaller overall oscillation.}
        \label{subfig:control-performance-comp}
  \end{subfigure}
  \caption{Comparison between the feasible command set, safe set and performance for the proposed BO-HZD controller with learned control parameters and HZD controller with manually tuned control parameters from~\cite{li2020animated} in a high-fidelity simulation.}
  \label{fig:manual-comparison}
  \vspace{-0.3cm}
\end{figure*}

\begin{figure*}
\centering
    \begin{subfigure}[t]{0.48\linewidth}
        \centering
        \includegraphics[width = 0.99\linewidth]{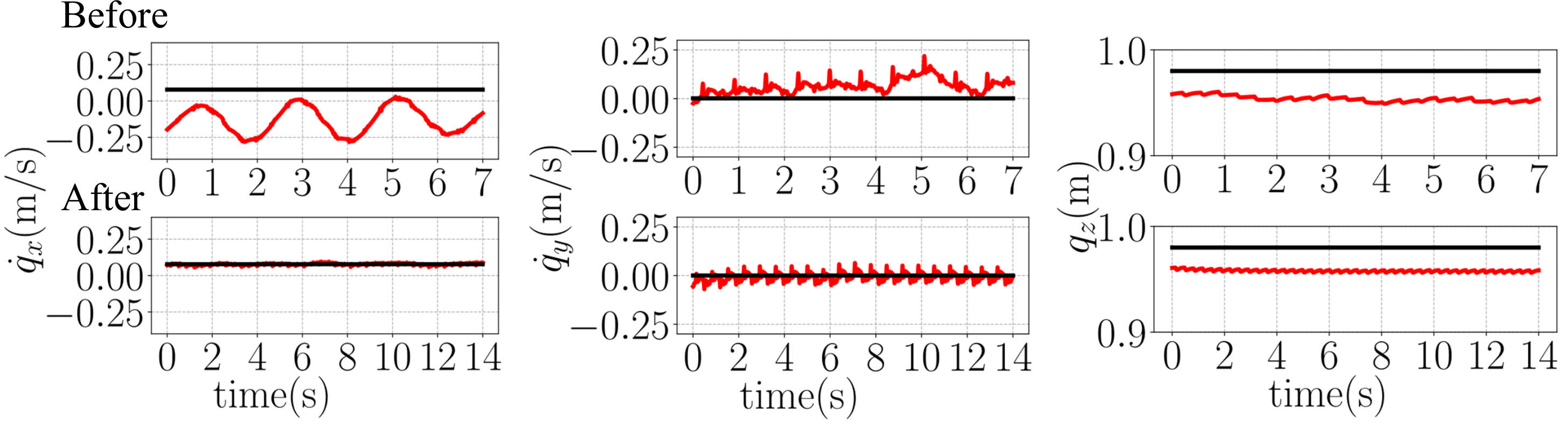}\\
        \caption{Normal walking height}
        \label{subfig:tracking1.00}
    \end{subfigure}
    \begin{subfigure}[t]{0.48\linewidth}
        \centering
        \includegraphics[width = 0.99\linewidth]{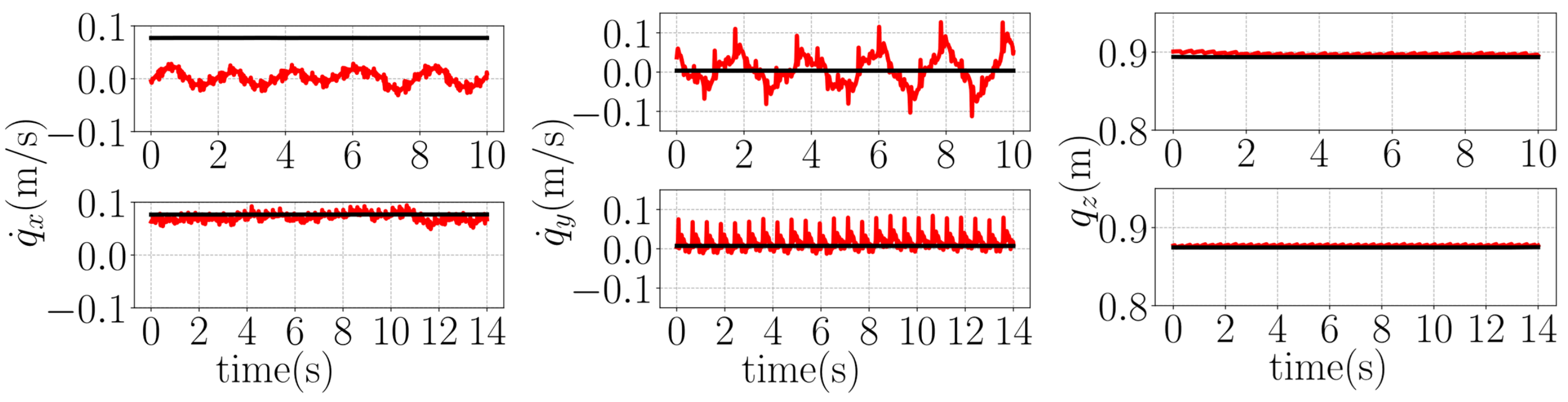}\\
        \caption{Medium walking height}
        \label{subfig:tracking0.85}
    \end{subfigure} \\
    \begin{subfigure}[t]{0.48\linewidth}
        \centering
        \includegraphics[width = 0.99\linewidth]{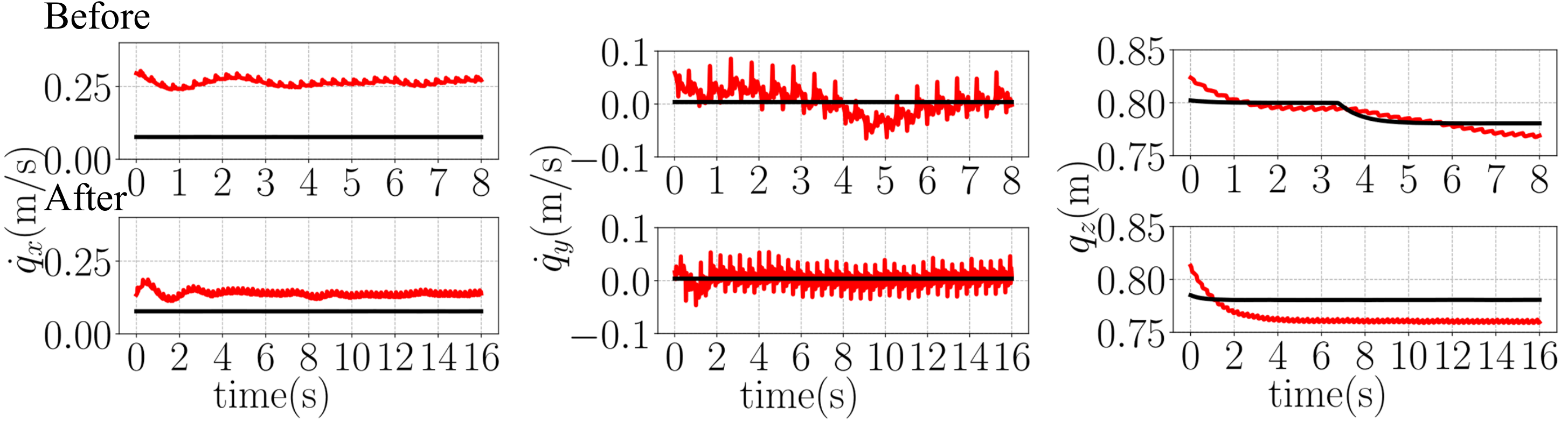}\\
        \caption{Low walking height}
        \label{subfig:tracking0.70}
    \end{subfigure}
    \begin{subfigure}[t]{0.48\linewidth}
        \centering
        \includegraphics[width = 0.99\linewidth]{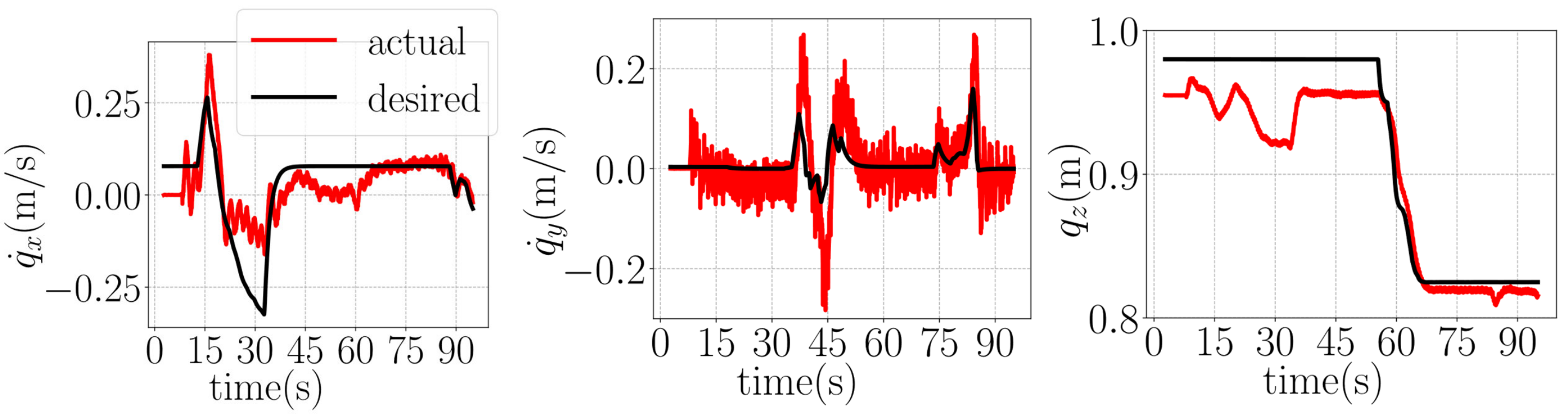}\\
        \caption{Controller performance at variable walking height}
        \label{subfig:tracking-all}
    \end{subfigure}
    \caption{{\bf Experimental Results}: Desired (black) and actual (red) traces are shown in the graph demonstrating comparisons of tracking errors and oscillations before and after real-world learning on Cassie for (a) a normal walking height $q_z$=1, (b) a medium walking height $q_z$ around 0.9, and (c) a low walking height $q_z$ around 0.8. The upper graphs in each subfigure depict the observed gait parameter $\hat{\mathbf{p}}$ when applying the parameters learned in the simulation directly onto the real-world robot and the lower graphs record $\hat{\mathbf{p}}$ after real-world learning. By using the real-world data to learn the sim-to-real discrepancy, both tracking error and oscillation are reduced. (d) The control performance using the proposed BO-HZD based controller at different heights with different saggital and lateral walking speeds.}
    \label{fig:tracking}
    \vspace{-0.4cm}
\end{figure*}

\subsection{Sample Efficiency}
Starting from random controller parameters, the entire learning process in simulation only takes 8000 iterations (one trajectory per iteration) to obtain optimal 6-dimensional PD gains for 310 different gait parameters which covers different saggital and lateral walking speeds, as well as different walking heights for Cassie. 
It only takes 30 iterations in the real world to learn the 6 dimensional corrections of PD gains $\bm{\delta}_{\mathbf{k}}$ and command tracking errors $\bm{\delta}_{\mathbf{p}}$ from simulation to the real world for different walking heights. 
The data collection for one iteration of the real-world optimization can be done in less than a minute. 
Compared to a previous RL method~\cite{li2021reinforcement} that obtains a controller for the same task of variable walking height control on Cassie and takes 400 million samples across more than 160 thousand trajectories to learn from scratch, this proposed method shows advantages of sampling efficiency.
Because of such sampling efficiency, we are able to apply the learning process in the high-fidelity simulator (rather than in Mujoco as done in the case of RL) whose computing time is slower than the real-time but brings us reliable control parameters.
However, it must be noted that BO is optimizing for the controller gains of a model-based method while RL obtains joint angle positions and is model-free.
 

\subsection{Evaluation in Simulation}
We benchmark the proposed HZD-based controller using the learned control parameter by Baysian optimization~(BO-HZD) with previous HZD-based controller with manually tuned control parameters from~\cite{li2020animated} whose parameters are in part based on~\cite{gong2019feedback} in the high-fidelity simulation on MATLAB SimMechanics.

One way to systematically compare control performance of different controllers on bipedal robots is by investigating the feasible command set and safe set, as introduced in detail in~\cite[Sec.IV]{li2021reinforcement}.
A controller is better if it has a larger feasible command set and safe set. The comparison of these two sets between the BO-HZD controller and HZD controller~\cite{li2020animated} are shown in Fig.~\ref{subfig:set-comp}. 
Quantitatively, BO-HZD results in 1656 feasible commands while HZD only has 724, which shows that BO-HZD can cover twice as large range of commands than HZD. 
Although the resulting safe sets shows a similar coverage in the converged gait parameter $\hat{\mathbf{p}}^\text{c}$ space between BO-HZD and HZD, BO-HZD has better tracking performance.
The average tracking error of BO is $|\mathbf{p}^\text{d}-\hat{\mathbf{p}}^\text{c}|=[0.0498,0.0395,0.0185]^T$ while the tracking error of HZD is $|\mathbf{p}^\text{d}-\hat{\mathbf{p}}^\text{c}|=[0.0919,0.0674,0.0445]^T$.
Such an advantage is also exemplified in Fig.~\ref{subfig:control-performance-comp} which shows the trajectory of $\mathbf{p}^\text{d}$ and $\hat{\mathbf{p}}^\text{c}$ of Cassie by using BO-HZD and HZD. 
Moreover, as shown in Fig.~\ref{subfig:control-performance-comp}, BO-HZD leads to less oscillation in saggital and lateral walking speeds of Cassie than HZD, which results in a smoother gait.


\subsection{Evaluation in Real World}
We further evaluate the proposed method in real-world experiments to control Cassie to track different desired gait parameters. Experiments are demonstrated in the video\footnote{\url{https://youtu.be/WxkdJdMRdfM}} and analyzed below.

Directly applying the simulation-learned controller results in lower than expected performance, as shown in top rows (labeled ``Before") of Figs.~\ref{subfig:tracking1.00}, \ref{subfig:tracking0.85}, \ref{subfig:tracking0.70} with different walking heights, illustrating the discrepancy between the simulation and real world due to the unmodeled dynamics. 
However, after using the real-world data to optimize for the corrections $\bm{\delta}_{\mathbf{k}}$ and $\bm{\delta}_{\mathbf{p}}$ for the PD gains and command tracking offsets, the tracking errors and oscillations can be significantly reduced as demonstrated in the bottom rows (labeled ``After") of Figs.~\ref{subfig:tracking1.00}, \ref{subfig:tracking0.85}, \ref{subfig:tracking0.70}.
Moreover, as we enforce the safety constraint for the gait stability for the robot, the controller never results in extreme unstable gaits in the real world. 

Furthermore, after the entire control parameter learning procedure has been applied, the resulting controller is able to successfully track varying desired gait parameters with small tracking errors and oscillation, as presented in Fig.~\ref{subfig:tracking-all}.

\section{Conclusion}
\label{sec:conclusion}
We have demonstrated a control parameter learning framework that applies BO to a HZD-based controller for achieving versatile locomotion control of bipedal robots. 
We leverage a multi-domain learning process that is firstly applied in a high-fidelity simulation to learn control parameters for different gaits and then in the real-world domain to tackle the sim-to-real gap by learning correction terms for the gains from simulation with safety constraints being enforced to maintain gait stability.
The learned controller was validated both numerically and experimentally on a person-sized bipedal robot Cassie, demonstrating significant improvement in terms of smaller steady-state tracking errors and smoother walking gaits in a wider range of walking motions compared to a previous HZD-based controller whose parameters were manually tuned.
By using BO-HZD, the controller is able to efficiently learn to gain schedule optimal control parameters for different walking gaits from a set of over 1800 control parameters for all the different gaits. This would have been impractical to tune by hand. 
As future work, we envision developing a sample efficient learning method that can utilize not only the state observation from the robot but also user preferences such as emotions of gaits.

\section*{Acknowledgements} 
This work was supported in part through National Science Foundation Grants CMMI-1944722 and CMMI-1931853. The authors would also like to thank Bike Zhang and Ayush Agrawal for their gracious help in experiments. 

\balance
\bibliographystyle{IEEEtran}
\bibliography{references}

\begin{thebibliography}{10}
\providecommand{\url}[1]{#1}
\csname url@rmstyle\endcsname
\providecommand{\newblock}{\relax}
\providecommand{\bibinfo}[2]{#2}
\providecommand\BIBentrySTDinterwordspacing{\spaceskip=0pt\relax}
\providecommand\BIBentryALTinterwordstretchfactor{4}
\providecommand\BIBentryALTinterwordspacing{\spaceskip=\fontdimen2\font plus
\BIBentryALTinterwordstretchfactor\fontdimen3\font minus
  \fontdimen4\font\relax}
\providecommand\BIBforeignlanguage[2]{{%
\expandafter\ifx\csname l@#1\endcsname\relax
\typeout{** WARNING: IEEEtran.bst: No hyphenation pattern has been}%
\typeout{** loaded for the language `#1'. Using the pattern for}%
\typeout{** the default language instead.}%
\else
\language=\csname l@#1\endcsname
\fi
#2}}

\bibitem{GrChAmSi2010}
J.~W. Grizzle, C.~Chevallereau, A.~Ames, and R.~Sinnet, ``3d bipedal robotic
  walking: Models, feedback control, and open problems,'' in \emph{IFAC
  Symposium on Nonlinear Control Systems}, 2010.

\bibitem{li2020animated}
Z.~Li, C.~Cummings, and K.~Sreenath, ``Animated cassie: A dynamic relatable
  robotic character,'' in \emph{2020 IEEE/RSJ International Conference on
  Intelligent Robots and Systems (IROS)}, 2020, pp. 3739--3746.

\bibitem{berkenkamp2021bayesian}
F.~Berkenkamp, A.~Krause, and A.~P. Schoellig, ``Bayesian optimization with
  safety constraints: safe and automatic parameter tuning in robotics,''
  \emph{Machine Learning}, pp. 1--35, 2021.

\bibitem{li2021reinforcement}
Z.~Li, X.~Cheng, X.~B. Peng, P.~Abbeel, S.~Levine, G.~Berseth, and K.~Sreenath,
  ``Reinforcement learning for robust parameterized locomotion control of
  bipedal robots,'' in \emph{International Conference on Robotics and
  Automation (ICRA)}, 2021, pp. 2811--2817.

\bibitem{xie2020learning}
Z.~Xie, P.~Clary, J.~Dao, P.~Morais, J.~Hurst, and M.~Panne, ``Learning
  locomotion skills for cassie: Iterative design and sim-to-real,'' in
  \emph{Conference on Robot Learning}.\hskip 1em plus 0.5em minus 0.4em\relax
  PMLR, 2020, pp. 317--329.

\bibitem{siekmann2021sim}
J.~Siekmann, Y.~Godse, A.~Fern, and J.~Hurst, ``Sim-to-real learning of all
  common bipedal gaits via periodic reward composition,'' in \emph{2021 IEEE
  International Conference on Robotics and Automation (ICRA)}.\hskip 1em plus
  0.5em minus 0.4em\relax IEEE, 2021, pp. 7309--7315.

\bibitem{castillo2021robust}
G.~A. Castillo, B.~Weng, W.~Zhang, and A.~Hereid, ``Robust feedback motion
  policy design using reinforcement learning on a 3d digit bipedal robot,'' in
  \emph{2021 IEEE/RSJ International Conference on Intelligent Robots and
  Systems (IROS)}.\hskip 1em plus 0.5em minus 0.4em\relax IEEE, 2021, pp.
  5136--5143.

\bibitem{hereid2017frost}
A.~Hereid and A.~D. Ames, ``Frost: Fast robot optimization and simulation
  toolkit,'' in \emph{2017 IEEE/RSJ International Conference on Intelligent
  Robots and Systems (IROS)}.\hskip 1em plus 0.5em minus 0.4em\relax IEEE,
  2017, pp. 719--726.

\bibitem{hereid2018rapid}
A.~Hereid, O.~Harib, R.~Hartley, Y.~Gong, and J.~W. Grizzle, ``Rapid trajectory
  optimization using c-frost with illustration on a cassie-series dynamic
  walking biped,'' pp. 4722--4729, 2019.

\bibitem{da20162d}
X.~Da, O.~Harib, R.~Hartley, B.~Griffin, and J.~W. Grizzle, ``From 2d design of
  underactuated bipedal gaits to 3d implementation: Walking with speed
  tracking,'' \emph{IEEE Access}, vol.~4, pp. 3469--3478, 2016.

\bibitem{gong2019feedback}
Y.~Gong, R.~Hartley, X.~Da, A.~Hereid, O.~Harib, J.-K. Huang, and J.~Grizzle,
  ``Feedback control of a cassie bipedal robot: Walking, standing, and riding a
  segway,'' in \emph{2019 American Control Conference (ACC)}, 2019, pp.
  4559--4566.

\bibitem{reher2021control}
J.~Reher and A.~D. Ames, ``Control lyapunov functions for compliant hybrid zero
  dynamic walking,'' \emph{arXiv preprint arXiv:2107.04241}, 2021.

\bibitem{WAFR2016GaitLibrarySteppingStones}
Q.~Nguyen, X.~Da, J.~W. Grizzle, and K.~Sreenath, ``Dynamic walking on stepping
  stones with gait library and control barrier,'' in \emph{Workshop on
  Algorithimic Foundations of Robotics (WAFR)}, 2016.

\bibitem{RSS2017DiscreteTerrainWalking}
Q.~Nguyen, X.~Da, W.~Martin, H.~Geyer, J.~W. Grizzle, and Sreenath, ``Dynamic
  walking on randomly-varying discrete terrain with one-step preview,'' in
  \emph{Robotics: Science and Systems (RSS)}, 2017.

\bibitem{rasmussen2003gaussian}
C.~E. Rasmussen, ``Gaussian processes in machine learning,'' in \emph{Summer
  school on machine learning}.\hskip 1em plus 0.5em minus 0.4em\relax Springer,
  2003, pp. 63--71.

\bibitem{DBLP:conf/icml/SrinivasKKS10}
\BIBentryALTinterwordspacing
N.~Srinivas, A.~Krause, S.~M. Kakade, and M.~W. Seeger, ``Gaussian process
  optimization in the bandit setting: No regret and experimental design,'' in
  \emph{Proceedings of the 27th International Conference on Machine Learning
  (ICML-10), June 21-24, 2010, Haifa, Israel}, J.~F{\"{u}}rnkranz and
  T.~Joachims, Eds.\hskip 1em plus 0.5em minus 0.4em\relax Omnipress, 2010, pp.
  1015--1022. [Online]. Available:
  \url{https://icml.cc/Conferences/2010/papers/422.pdf}
\BIBentrySTDinterwordspacing

\bibitem{bull2011convergence}
A.~D. Bull, ``Convergence rates of efficient global optimization algorithms.''
  \emph{Journal of Machine Learning Research}, vol.~12, no.~10, 2011.

\bibitem{jones2001taxonomy}
D.~R. Jones, ``A taxonomy of global optimization methods based on response
  surfaces,'' \emph{Journal of global optimization}, vol.~21, no.~4, pp.
  345--383, 2001.

\bibitem{mockus2012bayesian}
J.~Mockus, \emph{Bayesian approach to global optimization: theory and
  applications}.\hskip 1em plus 0.5em minus 0.4em\relax Springer Science \&
  Business Media, 2012, vol.~37.

\bibitem{lizotte2007automatic}
D.~J. Lizotte, T.~Wang, M.~H. Bowling, and D.~Schuurmans, ``Automatic gait
  optimization with gaussian process regression.'' in \emph{IJCAI}, vol.~7,
  2007, pp. 944--949.

\bibitem{calandra2016bayesian}
R.~Calandra, A.~Seyfarth, J.~Peters, and M.~P. Deisenroth, ``Bayesian
  optimization for learning gaits under uncertainty,'' \emph{Annals of
  Mathematics and Artificial Intelligence}, vol.~76, no.~1, pp. 5--23, 2016.

\bibitem{tucker2020preference}
M.~Tucker, E.~Novoseller, C.~Kann, Y.~Sui, Y.~Yue, J.~W. Burdick, and A.~D.
  Ames, ``Preference-based learning for exoskeleton gait optimization,'' in
  \emph{2020 IEEE International Conference on Robotics and Automation
  (ICRA)}.\hskip 1em plus 0.5em minus 0.4em\relax IEEE, 2020, pp. 2351--2357.

\bibitem{csomay2021learning}
N.~Csomay-Shanklin, M.~Tucker, M.~Dai, J.~Reher, and A.~D. Ames, ``Learning
  controller gains on bipedal walking robots via user preferences,''
  \emph{arXiv preprint arXiv:2102.13201}, 2021.

\bibitem{tesch2011using}
M.~Tesch, J.~Schneider, and H.~Choset, ``Using response surfaces and expected
  improvement to optimize snake robot gait parameters,'' in \emph{2011 IEEE/RSJ
  International Conference on Intelligent Robots and Systems}.\hskip 1em plus
  0.5em minus 0.4em\relax IEEE, 2011, pp. 1069--1074.

\bibitem{ryou2021multi}
G.~Ryou, E.~Tal, and S.~Karaman, ``Multi-fidelity black-box optimization for
  time-optimal quadrotor maneuvers,'' \emph{The International Journal of
  Robotics Research}, p. 02783649211033317, 2021.

\bibitem{10.5555/3020751.3020778}
M.~A. Gelbart, J.~Snoek, and R.~P. Adams, ``Bayesian optimization with unknown
  constraints,'' in \emph{Proceedings of the Thirtieth Conference on
  Uncertainty in Artificial Intelligence}, ser. UAI'14.\hskip 1em plus 0.5em
  minus 0.4em\relax Arlington, Virginia, USA: AUAI Press, 2014, p. 250–259.

\bibitem{sui2015safe}
Y.~Sui, A.~Gotovos, J.~Burdick, and A.~Krause, ``Safe exploration for
  optimization with gaussian processes,'' in \emph{International Conference on
  Machine Learning}.\hskip 1em plus 0.5em minus 0.4em\relax PMLR, 2015, pp.
  997--1005.

\bibitem{schreiter2015safe}
J.~Schreiter, D.~Nguyen-Tuong, M.~Eberts, B.~Bischoff, H.~Markert, and
  M.~Toussaint, ``Safe exploration for active learning with gaussian
  processes,'' in \emph{Joint European conference on machine learning and
  knowledge discovery in databases}.\hskip 1em plus 0.5em minus 0.4em\relax
  Springer, 2015, pp. 133--149.

\bibitem{berkenkamp2016safe}
F.~Berkenkamp, A.~P. Schoellig, and A.~Krause, ``Safe controller optimization
  for quadrotors with gaussian processes,'' in \emph{2016 IEEE International
  Conference on Robotics and Automation (ICRA)}, 2016, pp. 491--496.

\bibitem{marco2021robot}
A.~Marco, D.~Baumann, M.~Khadiv, P.~Hennig, L.~Righetti, and S.~Trimpe, ``Robot
  learning with crash constraints,'' \emph{IEEE Robotics and Automation
  Letters}, 2021.

\bibitem{frazier2018tutorial}
P.~I. Frazier, ``A tutorial on bayesian optimization,'' \emph{stat}, vol. 1050,
  p.~8, 2018.

\bibitem{li2021vision}
Z.~Li, J.~Zeng, S.~Chen, and K.~Sreenath, ``Vision-aided autonomous navigation
  of underactuated bipedal robots in height-constrained environments,''
  \emph{arXiv preprint arXiv:2109.05714}, 2021.

\end{thebibliography}

\end{document}